\documentclass{article}

\usepackage{arxiv}
\usepackage{amsmath}

\usepackage[utf8]{inputenc} 
\usepackage[T1]{fontenc}    
\usepackage{hyperref}       
\usepackage{url}            
\usepackage{booktabs}       
\usepackage{amsfonts}       
\usepackage{nicefrac}       
\usepackage{microtype}      
\usepackage{lipsum}
\usepackage{graphicx}

\title{Identifying Mislabeled Images in Supervised Learning Utilizing Autoencoder}

\author{
  Yunhao Yang \\
  Department of Computer Science\\
  University of Texas at Austin\\
  Austin, TX 78705 \\
  \texttt{yunhaoyang234@utexas.edu} \\
  
  \AND
  Andrew Whinston \\
  Department of Information, Risk and Operations Management
  \\ Department of Computer Science \\
  University of Texas at Austin\\
  Austin, TX 78705 \\
  \texttt{abw@utexas.edu} \\
}

\begin{document}
\maketitle

\begin{abstract}
Supervised learning is based on the assumption that the ground truth in the training data is accurate. However, this may not be guaranteed in real-world settings. Inaccurate training data will result in some unexpected predictions. In image classification, incorrect labels may cause the classification model to be inaccurate as well. In this paper, I am going to apply unsupervised techniques on the training data before training the classification network. A convolutional autoencoder is applied to encode and reconstruct images. The encoder will project the image data on to latent space. In the latent space, image features are preserved in a lower dimension. The assumption is that data samples with similar features are likely to have the same label. Noised samples can be classified in the latent space by the Density-Base Scan (DBSCAN) clustering algorithm. These incorrectly labeled data are visualized as outliers in the latent space. Therefore, the outliers identified by the DBSCAN algorithm can be classified as incorrectly labeled samples. After the outliers are detected, all the outliers are treated as mislabeled data samples and removed from the dataset. Thus the training data can be directly used in training the supervised learning network. The algorithm can detect and remove above 67\% of mislabeled data in the experimental dataset.
\end{abstract}

\keywords{Supervised Learning \and Noised Label \and Autoencoder \and Image Classification \and DBSCAN \and Outlier Detection \and Cluster \and Pre-processing}

\section{Introduction}
Supervised learning is one of the machine learning techniques that maps input to output based on given input-output pairs in the training data. The training data for supervised learning consists of a set of labeled samples. Each labeled sample consists of input data and the desired output that corresponding to the input data. The goal of supervised learning is to analyze the training data and produce an inferred function, which can be used for mapping new examples. However, supervised learning is based on the assumption that all the labels in the training data are accurate. Unlike reinforcement learning and unsupervised learning, data labeling plays an important role. When the accuracy of training data is taken into consideration, a way of adjusting or removal of inaccurate labels is necessary.

Collecting data from experts is common in supervised learning. For instance, in the medical area, labeling requires people who have expertise such as medical doctors to make judgments on the disease. Doctors generally have a certain malpractice rate. Therefore, it is important to evaluate their judgments and reduce the misjudged data in the training data to reduce the malpractice rate of our trained model.

In this paper, I will present a straight forward approach for classifying mislabeled training samples. A convolutional autoencoder is utilized for the classification. Autoencoder is a type of artificial neural network that trains data in a supervised manner. The convolutional autoencoder is one of the most effective and commonly used autoencoders, which consists of multiple convolutional layers. In the context of a convolutional neural network, convolution is a linear operation that involves the multiplication of a set of weights with the input, much like a traditional neural network. Given that the technique was designed for two-dimensional input, the multiplication is performed between an array of input data and a two-dimensional array of weights, called a filter or a kernel\cite{convlayer}.

The encoding procedure performs a dimension reduction to training data and projects training data on the latent space, where the features of training data are preserved. A decoder can be constructed to recover the data using the features in the latent space. The quality of the reconstruction is depended on the quality of features in the latent space and the complexity of the network. Therefore, in this paper, I will measure the reconstruction quality to evaluate the quality of latent space projections.

Latent space is helpful for learning data features and for obtaining simpler representations of data for analysis. It is a representation of dimension-reduced data in which data points fall in the same category are closer together in space. Therefore, if a clustering algorithm is applied in the latent space, the data points in the same category are more likely to be in the same cluster. This paper is based on the assumption that if a specific data point falls in a cluster in which the majority of data points in this cluster consist of a label other than the chosen data point, then this chosen data point is classified as mislabeled data.

\section{Related Work}
There are several approaches to minimize the negative effect of inaccurate training labels. Adjusting loss function using label noise statistics is a commonly known methodology. A large number of researchers retain the network architecture, training data, and training procedures while only modifying the loss function. In \cite{9000595}, an automated methodological framework can be proposed to identify mislabeled data using two metric functions. Cross-entropy Loss that indicates divergence between a prediction and ground truth, and Influence function that reflects the dependence of a model on data.

Another common approach is to adjust the network architecture. Several studies have proposed attaching a “de-noise layer” to the end of their deep learning networks. The de-noise layer proposed by \cite{sukhbaatar2014training} is equivalent to multiplication with the transition matrix between noisy and true labels. The authors developed methods for learning this matrix in parallel with the network weights using error back-propagation. A similar noise layer was proposed by \cite{thekumparampil2018robustness} for training a generative adversarial network (GAN) under label noise. \cite{song2020learning} proposed methods for estimating the transition matrix from either a clean or a noisy dataset\cite{karimi2019deep}. Besides adding a de-noise layer, \cite{yao2017deep} proposed a model that identifies the mismatch between the latent and noisy labels by embedding the quality variables into different sub-spaces, which effectively minimizes the noise effect. These studies modify the network architecture to identify noised labels and minimize the negative effect on the noised labels.

Modifying the training scheme while preserving network architecture is also an effective approach to eliminating or reducing noised training samples. Mixup\cite{zhang2017mixup} is a less intuitive but simple and effective learning principle to alleviate the impact of inaccurate labeling. It trains a neural network on convex combinations of pairs of examples and their labels. By doing so, mixup regularizes the neural network to favor simple linear behavior in-between training examples.

This paper is following another approach, which is exploiting data-label consistency to identifying and eliminating inaccurate labels in the pre-processing stage. The noised samples are expected to be removed or reclassified prior to training the network. This is a straightforward and efficient approach in which the de-noising procedure is independent of the overall training process. Because an unsupervised learning technique is applied explicitly prior to the training process, while no labeling is required at this stage
\cite{yang2020survey}
. This means the de-noising model can be combined with any kind of model without modifying the structure of the model. \cite{autoencoder} presented a way of using autoencoder to project data to the latent space and remove outliers of specific classes. Furthermore, they proposed a weighted class-specific autoencoder by considering the effect of each data point on the postulated model. The paper Robust Class-Specific Autoencoder for Data Cleaning and Classification in the Presence of Label Noise\cite{autoencoder} shows that in Class-Specific Autoencoder, the reconstruction error of the noised dataset is significantly higher than the same dataset without noise. Which indicates the noise samples does negatively affect the neural network training procedure. This paper also proves that a mislabeled data sample is more likely to be identified as an outlier and assigned to a lower weight.

\section{Methodology}
\subsection{Network Architecture}
This paper is concentrating on noise removal in the pre-processing stage, where the incorrectly labeled samples are detected and removed according to a specified methodology. A convolutional autoencoder is applied to the training data in advance to utilize unsupervised techniques to detect mislabeled samples.

\begin{figure}[!htbp]
\begin{center}
    \includegraphics[width=0.9\linewidth]{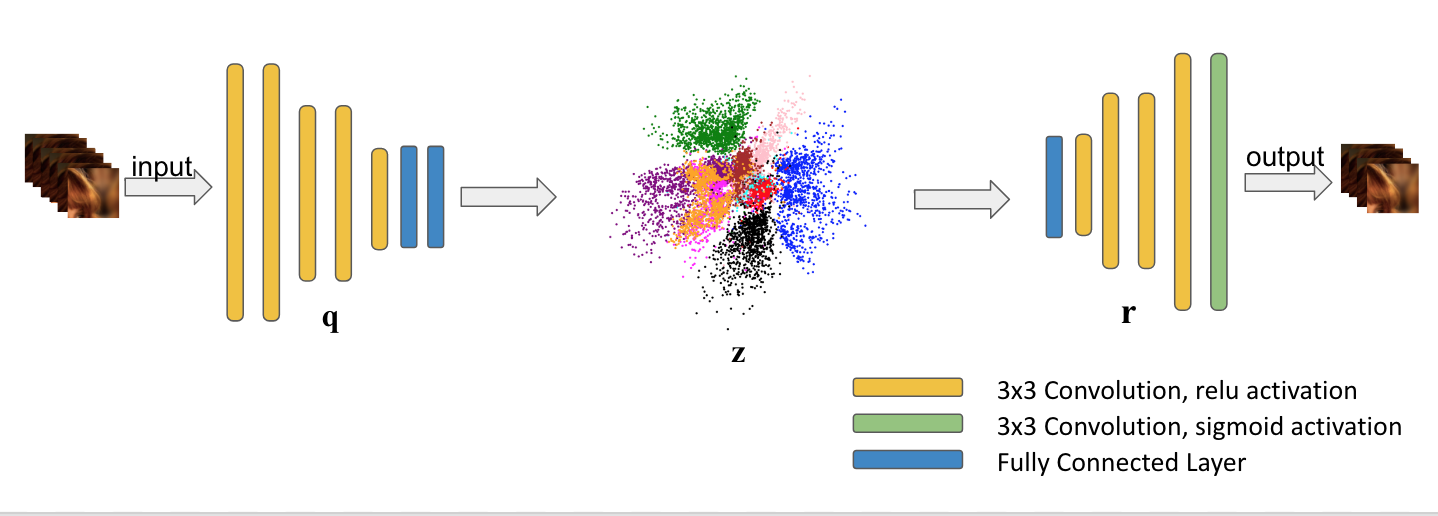}
\end{center}
\caption{Network Architecture: the network is consists of a simple convolutional autoencoder. The encoder is composed of five $3\times 3$ kernel convolutional layers and two fully connected layers sequentially. The decoder consists of one fully connected layer and five convolutional transpose layers, which are symmetric to the convolutional layers in the encoder.}
\end{figure}

The convolution autoencoder consists of two primary components, the encoder, and the decoder. Within the encoding stage, images are encoded and projected into the latent space. This reduces the dimension of the images while preserving the primary features of these images. The assumption is that images with identical labels presumably contain similar features. Therefore, the euclidean distance between the projections contains the same label tend to be relatively shorter. Then, a clustering algorithm is applied to the latent space to separate the images with different features. Images in the same cluster are more likely to consist the same label. Thus we can classify incorrectly labeled data.

The decoder reconstructs images from the features preserved in the latent space. The reconstruction quality is positively related to the quality of feature projection. The reconstructed images are evaluated using Peak signal-to-noise ratio (PSNR) \cite{5596999}.
\begin{equation}
\label{eqn:psnr}
  \mbox{PSNR} = 10\log_{10} \frac{(2^d-1)^2WH}{\sum_{i=1}^W \sum_{j=1}^H (p[i,j]-p'[i,j])^2} 
\end{equation}

\begin{align*}
& W \quad \text{image width} \\
& H \quad \text{image height} \\
& p \quad \text{original image} \\
& p'\quad \text{reconstructed image}
\end{align*}
The accuracy of the classification on incorrect labels is the ratio of the number of correct labels over the original training data size. The quality of reconstructions will be compared with the accuracy of identifying the incorrect labels. We are expecting a positive relation between these two metrics.

\subsection{Loss Function}
The loss functions for training the convolutional autoencoder are mean square error and Kullback–Leibler divergence (KL Loss) score\cite{Joyce2011}. The mean square error is used to compute the variance between the reconstructed image and the source image. The KL loss is a measure of how the probability distribution of reconstructions is different from the distribution of the source data. The equations are shown below.
\begin{equation}
    \mathcal{L}= \beta_{KL}\mathcal{L}_{KL} + \mathcal{L}_{MSE}
\end{equation}
\begin{equation}
    \mathcal{L}_{KL}= \sum_{i=1}^N \frac{1}{2}\left[ \sum_{j} (1 +\log \vec{\sigma}_j^2(\vec{x}_i|\theta))+\|\vec{\mu}(\vec{x}_i|\theta)\|_2^2 +\|\vec{\sigma}(\vec{x}_i|\theta)\|_2^2 \right]
\end{equation}
\begin{equation}
    \mathcal{L}_{MSE}=  \sum_{i=1}^N\|\hat{\vec{x}}_i -\vec{x}_i\|_2^2
\end{equation}

Here $\beta_{KL}$ is the parameter that controls the total loss function. N is the batch size of the input. $\vec{\mu}$ and $\vec{\sigma}$ are the mean and variance computed from the training samples within each batch.

\subsection{Noise Detection and Removal}
\label{section:noise}
This paper presents one of the common methodologies approaching noise detection: Density-Based Scan Clustering (DBSCAN)\cite{dbscan}.

DBSCAN is relying on a density-based notion of clusters which is designed to discover clusters of arbitrary shape. DBSCAN requires only one input parameter and supports the user in determining an appropriate value for it\cite{dbnoise}. DBSCAN algorithm requires 2 parameters- epsilon, which specifies how close points should be to each other to be considered a part of a cluster; and minimum points, which specifies how many neighbors a point should have to be included into a cluster. Therefore, all the points that are not assigned to any clusters are considered outliers.

\begin{figure}[!htbp]
\begin{center}
    \includegraphics[width=0.9\linewidth]{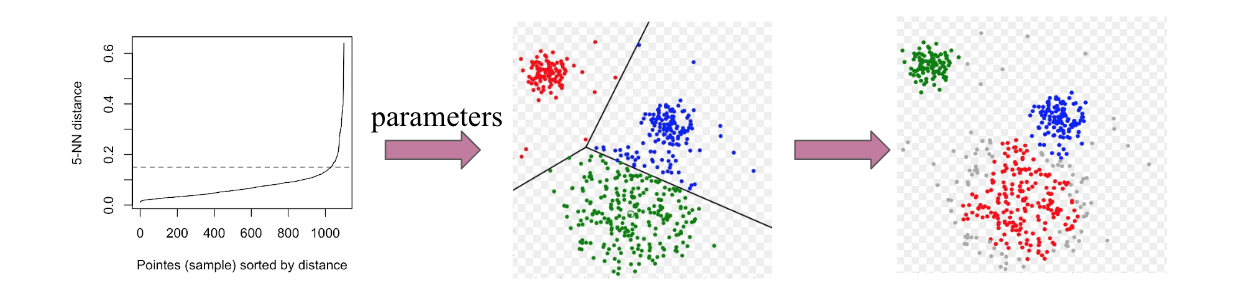}
\end{center}
\caption{Outlier Detection: after the latent space projections are retrieved, for each class, Rahmah's algorithm is applied to determine the best parameters for the DBSCAN algorithm. Then, the projects are separated by classes and the DBSCAN clustering algorithm is applied to each class of samples respectively to identify outliers.}
\end{figure}

The DBSCAN clustering algorithm is applied to each class in the training data. By assumption, the samples consist of the same class label most likely will contain identical features. So, while applying the DBSCAN clustering algorithm to all the samples belong to a chosen class, these encoded samples in the latent space are supposed to be classified in one cluster. Thus the outliers identified by the DBSCAN algorithm have a large possibility to be the mislabeled samples. These outliers will be removed from the training data.

In order to determine the hyper-parameter- epsilon, Rahmah's algorithm\cite{Rahmah_2016} is applied to the samples that consist of the same label. The algorithm is applied to one label class at a time. This means we need to determine and record 10 sets of parameters corresponding to 10 classes in a dataset. The algorithm begins with the calculation of Euclidean distance on each pair of data. A function of the matrix is used because the results dist upper triangular form that needs to be normalized into the matrix intact. Normalization is used to facilitate the search for the k nearest neighbors on each line of the distance calculation.

After calculating the distance and the matrix, further searches carried out k nearest neighbors of each matrix line spacing for sorting is done in ascending for each result the closest distance to the neighbors. Sorting the results of each line of the nearest neighbor made a plot with the x-axis and the y-axis is the object and the distance k nearest neighbors. Plots that have formed in ascending calculated the difference in slope of the line to get the value of epsilon. A point that has the most significant slope changes will be the optimal value for epsilon.

\section{Experiment}

\subsection{Dataset}

Modified National Institute of Standards and Technology database (MNIST)\cite{lecun2010mnist}, Fashion-MNIST, and Canadian Institute For Advanced Research (Cifar) 10\cite{Krizhevsky09learningmultiple} are the datasets used in the experiments. MNIST is a large database of handwritten digits that is commonly used for training various image processing systems. There are 10 classes in the MNIST dataset (digits from 0 to 9). It contains 60,000 $28\times 28$ resolution training images and 10,000 testing images. Cifar 10 is a dataset of 50,000 $32\times 32$ color training images and 10,000 test images, identified over 10 categories. Fashion-MNIST\cite{xiao2017fashionmnist} is a dataset of Zalando’s article images consisting of a training set of 60,000 examples and a test set of 10,000 examples. Each example is a $28\times 28$ grayscale image, associated with a label from 10 classes. Fashion-MNIST is intended to serve as a direct drop-in replacement of the original MNIST dataset for benchmarking machine learning algorithms.

All the datasets contain images in various classes and each image in the dataset associates with a categorical label. Images with the identical label have their unique characteristics that can be differentiated from images in other categories. All three datasets are commonly used in various experiments because of their precise classification and relatively small data scale. Therefore, the outcomes observed from the datasets are convincing.

\begin{figure}[!htbp]
\begin{center}
    \includegraphics[height=36mm]{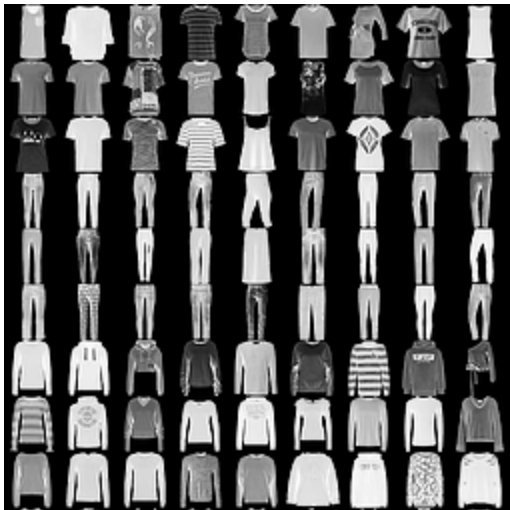}
    \includegraphics[height=36mm]{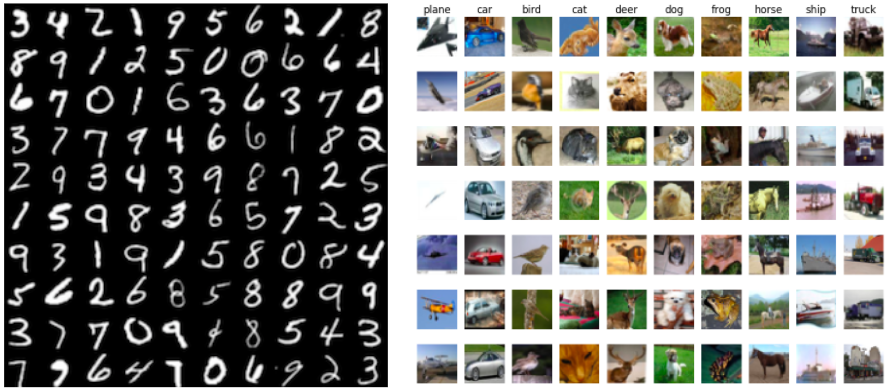}
\end{center}
\caption{Left figure is Fashion-MNIST. Middle figure is Modified National Institute of Standards and Technology database (MNIST). Right figure is Canadian Institute For Advanced Research (Cifar) 10}
\end{figure}

\subsection{Noise Generation}
At the beginning of the experiment, noised data are generated from the MNIST, Fashion-MNIST, and Cifar datasets. For each dataset, a certain percent of training samples are randomly selected. The selected samples are assigned to a random class label other than its own label. In this experiment, the noised dataset consists of $15\%$ mislabeled samples and $85\%$ accurate samples.

\subsection{Experiment Scheme}
The images in the noise dataset are utilized to train the convolutional autoencoder. Each image $\mathcal{X}$ is the input of the encoder, as well as the target of reconstruction. The reconstruction loss of the convolutional autoencoder is computed by
\begin{center}
    $\mathcal{L}_{recons} = \mathcal{L}_{MSE} (\mathcal{X}, \phi (\mathcal{X}))$
\end{center}
where $\phi (\mathcal{X})$ refers to the reconstruction of image $\mathcal{X}$ that generated by the encoder and decoder. The encoder consists of five convolutional layers, one flatten layer and two dense layers. The decoder consists of one dense layer and six convolutional layers. Each layer associates with a kernel regularizer, which performs both L1 and L2 Regularization on the layer to avoid overfitting. Each encoder has approximately  450 thousand parameters, and each decoder has 400 thousand parameters. The model is trained on a single 12GB NVIDIA Tesla K80 GPU with a batch size of 128. The parameters are optimized in 10 epochs using the Adam optimizer with $\beta_1 = 0.9$ and $\beta_2 = 0.999$. The average PSNR score is computed base on the reconstructed images and original images, the score is recorded after the training procedure. An experiment that analyze the correlation between PSNR and denoising performance is conducted.

After finish training the autoencoder, each image is encoded and projected onto the latent space by the encoder. The noise-detection methodology is applied to classes in the latent space one at a time. First, a certain class label is selected and all the samples consist of the selected class label are extracted from the dataset. Second, the methodology presented in \ref{section:noise} is utilized to determine the best hyper-parameter to use in the DBSCAN clustering algorithm. Then, the DBSCAN algorithm is applied to classify outliers. These outliers are considered mislabeled samples and removed from the dataset. Finally, after the removal of outliers from all the classes, the remaining training samples, and their labels are evaluated by comparing them with the correct labels.

\subsection{Evaluation}
In the experiment, the labels in the reclassified training dataset are compared with the correct labels from the original dataset. The accuracy of the reclassified dataset is compared with the noised dataset that is generated in the experiment. The accuracy is evaluated by the Jaccard index:

\begin{equation}
    \mathcal{J}_{\mu}(D) = \frac{\mu (S \cap D)}{\mu (S \cup D)}
\end{equation}

where \textit{S} is the set of labels from the source dataset. $\mu$ is a measure on a measurable space X. In this scenario, $\mu$ measures the cardinality of the given set. The performance of the convolutional autoencoder is measured by
\begin{center}
    $\mathcal{P} = \mathcal{J}_{\mu}(D) -  \mathcal{J}_{\mu}(N) $
\end{center}
where \textit{D} and \textit{N} are the sets of labels from the denoised dataset and noised dataset respectively. A higher $\mathcal{P}$ represents a more significant enhancement on the Jaccard index, which indicates the better performance of the convolutional autoencoder. Conversely, a negative $\mathcal{P}$ indicates the convolutional autoencoder generates more noised samples during the denoising procedure.

\section{Discussion}
The convolutional autoencoder is trained for reconstructing the images, while the transformation to latent space is performed by the encoder. The model is to learn the ‘structural similarities’ between images. In fact, this is how the model is able to classify labels in the first place- by learning the features of each image. Because the model is required to then reconstruct the compressed data, it must learn to store all relevant information and disregard the noise. This is the value of compression- it allows us to get rid of any extraneous information, and only focus on the most important features. The figure below shows the reconstructions of the samples generated from the latent space projections by the decoder.

\begin{figure}[!htbp]
\begin{center}
    \includegraphics[width=0.9\linewidth]{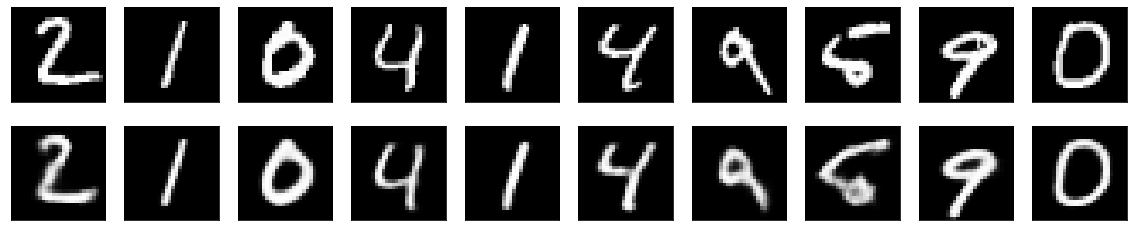}
    \includegraphics[width=0.9\linewidth]{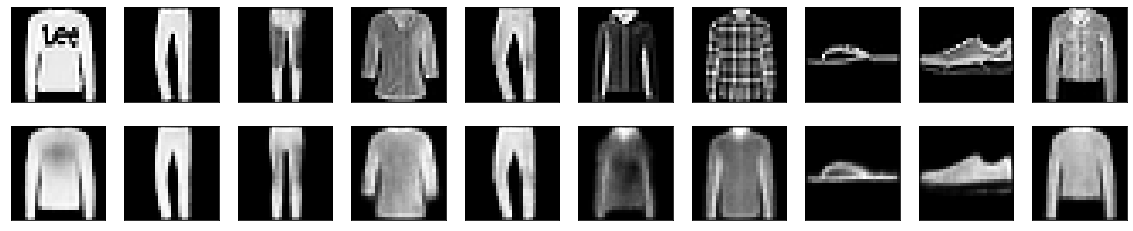}
\end{center}
\caption{Reconstruction results: the top two rows are the original samples and reconstructions from MNIST. The bottom two rows are the original samples and reconstructions from Fashion-MNIST.}
\end{figure}

The convolutional autoencoder preserves the overall structure and significant features while compressing the data. Therefore, some details may be lost during the encoding and decoding procedure. However, this means the variance of samples belong to the same class is smaller compared to the original samples. The example from Fashion-MNIST is more convincing, the decorations of the clothes are disappeared in the reconstructions. This is beneficial for clustering the samples.

The latent space dimension I selected in the experiment is 24, which gives the best latent space visualization. To visualize the latent space, I use Principle Component Analysis\cite{pca} to reduce the dimension of latent space to two-dimensional space. Principal component analysis (PCA) is the process of computing the principal components and using them to perform a change of basis on the data, sometimes using only the first few principal components and ignoring the rest. The first two principal components are selected for plotting the data point projections in the latent space.

\begin{figure}[!htbp]
\begin{center}
    \includegraphics[width=0.45\linewidth]{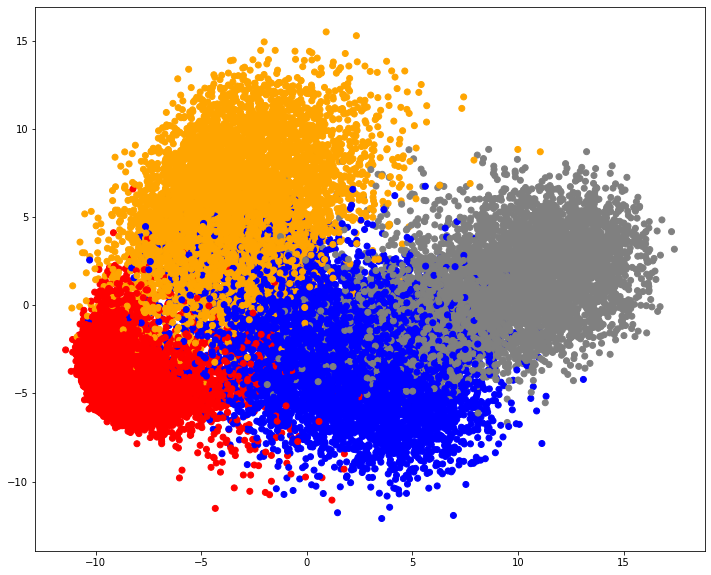}
    \includegraphics[width=0.45\linewidth]{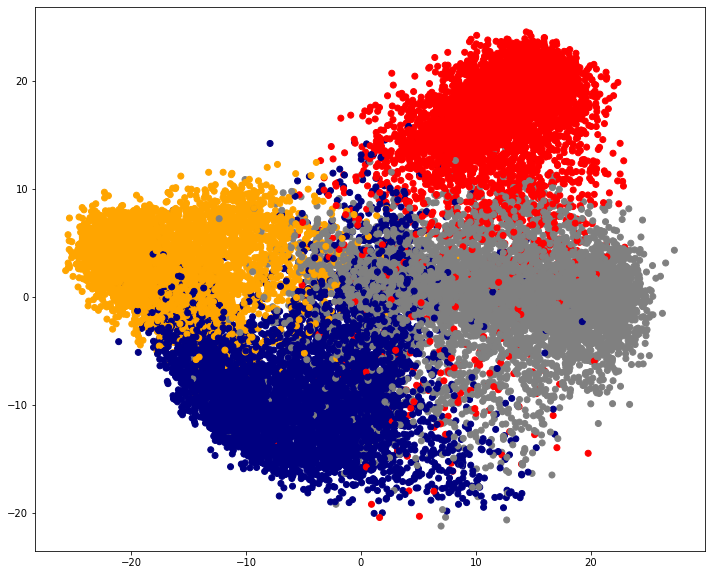}
\end{center}
\caption{Latent space projections of selected samples in four different classes from MNIST dataset (left) and Fashion-MNIST (right).}
\end{figure}

The figure indicates training samples from different classes (consisting of different labels) are separated. Therefore, the assumption of this paper holds: training samples from the same class are likely to be assigned to the same cluster while a clustering algorithm is applied. Then, the DBSCAN clustering algorithm is used to identify and remove noise training samples. 

\begin{figure}[!htbp]
\begin{center}
    \includegraphics[width=0.45\linewidth]{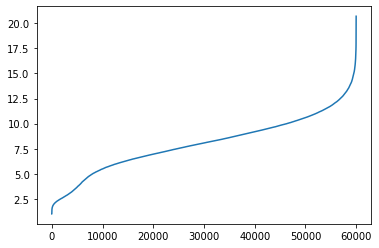}
\end{center}
\caption{Distances to Nearest Neighbor (MNIST): this shows an abstract view of how to get the parameter $\epsilon$ for the DBSCAN algorithm. In this case, $\epsilon = 10$ is selected. However, in the real experiments, the parameter $\epsilon$ is determined separately for each class, thus different class in the dataset may associate with different parameter $\epsilon$.}
\label{fig:db}
\end{figure}

In order to determine parameter $\epsilon$ in the DBSCAN algorithm, I use Rahmah's algorithm to plot the distances between each point to its nearest neighbor. Figure \ref{fig:db} shows an example for the MNIST dataset, a point in the plot that has the most significant slope changes is the optimal value for $\epsilon$. In the experiment, I choose $\epsilon = 10$ for all the classes in the MNIST dataset since the noise labels are randomly assigned, thus each class consists of the same proportion of noise samples. Then, I fix the minimum number of samples within $\epsilon$ equals 5 for the MNIST dataset. The $\epsilon$ for the other two datasets- Fashion MNIST and Cifar 10 are determined by the same methodology, while the minimum number of samples within $\epsilon$ is fixed to 5. The $\epsilon$ for Fashion-MNIST and Cifar 10 is 8 and 5 respectively.

Due to the nature of the noise generation in this experiment, each class consists of approximately the same proportion of noise samples. Therefore, a universal set of parameters for the DBSCAN algorithm can be determined. However, in reality, the noise may not be uniformly distributed to each class, there may exist noise imbalance between classes. Then, Rahmah's algorithm is required to be applied to one class at a time, rather than applied to the entire dataset.

\begin{figure}[!htbp]
\begin{center}
    \includegraphics[width=0.3\linewidth]{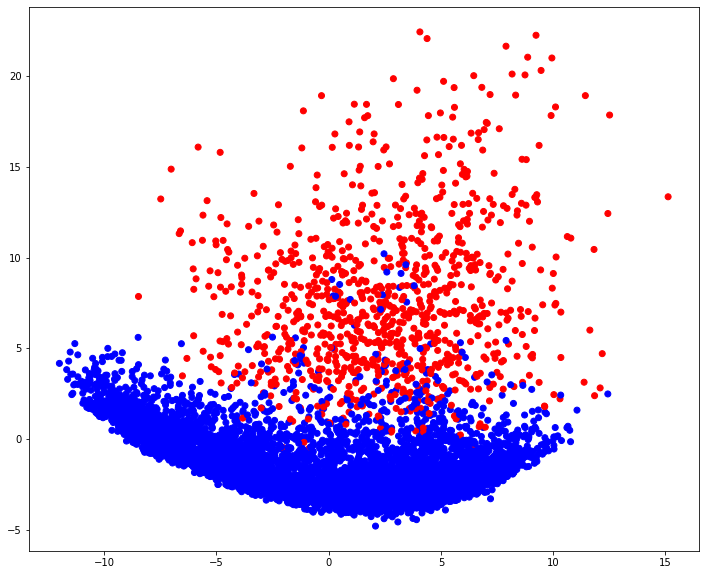}
    \includegraphics[width=0.3\linewidth]{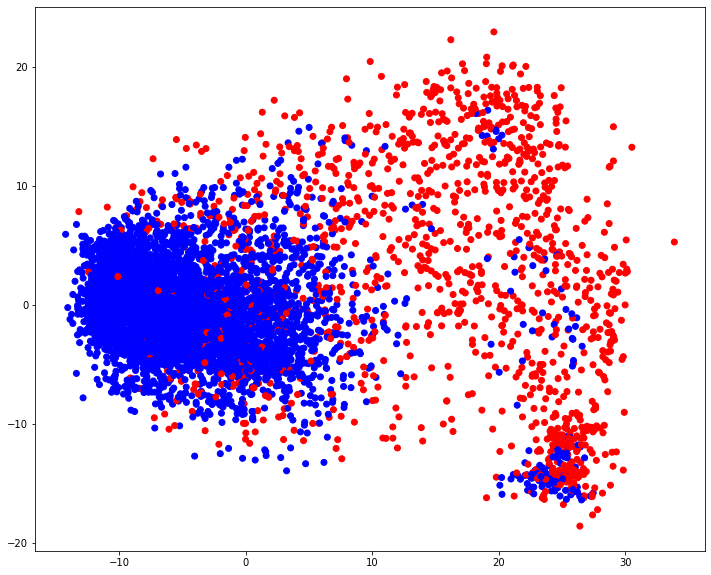}
    \includegraphics[width=0.3\linewidth]{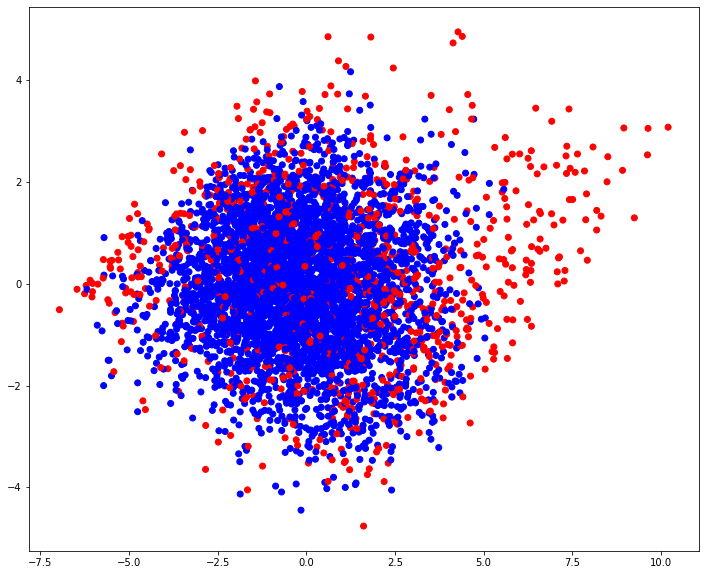}
\end{center}
\caption{DBSCAN Results: the left figure is a plot of samples in a selected class from MNIST; the middle figure is a plot of samples in a selected class from Fashion-MNIST; and the right figure is selected from Cifar 10. Blue points are the core and bordered points, red points are outliers.}
\label{fig:noise}
\end{figure}

The DBSCAN algorithm works as the following:
First, randomly select a point not already assigned to a cluster or designated as an outlier. Determine if it’s a core point by seeing if there are at least $M$ points around it within $\epsilon$ distance, where $M$ is the minimum number of samples within $\epsilon$ distance.
Second, create a cluster of this core point and all points within the $\epsilon$ distance of it.
Last, find all points that are within the $\epsilon$ distance of each point in the cluster and add them to the cluster. Find all points that are within the $\epsilon$ distance of all newly added points and add these to the cluster. Any point that is randomly selected that is not found to be a core point or a borderline point is classified as a noise point or outlier and is not assigned to any cluster. Thus, it does not contain at least $M$ points that are within $\epsilon$ distance from it or are not within $\epsilon$ distance of a core point\cite{jose}. Some selected DBSCAN noise classification examples are shown in Figure \ref{fig:noise}.

\subsection{Benchmark}
After the noise points are identified, all the noise points are removed from the training dataset. Then, the accuracy of the remaining training data are computed and the performance of the model is evaluated by the methodology stated in Section 4.4. In addition, there are three benchmark methods for removing noised training data. The three benchmarks are compared with our model.

\subsubsection{Benchmark 1: KNN on Image Vector}
Benchmark 1 applies K Nearest Neighbor (KNN) clustering algorithm directly on the source image vectors. First, all the source images are flattened to one dimensional vectors (for example, MNIST image is transformed from $28 \times 28$ to a vector of length 784). Then, a subset of training data which consists of 1,000 image-label pairs. is selected and serve as the representatives. All the training samples are classified by their K nearest representatives. The classification result is used to replace the original inaccurate class labels.

\subsubsection{Benchmark 2: Principle Component Analysis}
Benchmark 2 utilizes the eigenvectors of the training data. Prior to the training stage, a subset of training data that has 1,000 image-label pairs is selected and serve as the representatives. In the training stage, a set of eigenvectors is computed from the representatives and sorted by their eigenvalues. Then, images are projected on the eigenspace and classified using the k-nearest neighbor algorithm.

The representatives are stored in a matrix A. Eigenvector $v = [v_1, ..., v_n]$ and eigenvalue $\lambda$ for a matrix A satisfies the following:
\begin{center}
    $(A - \lambda I)v = 0$
\end{center}
Every eigenvalue is corresponding to a particular eigenvector. Each pair of eigenvectors are orthogonal to each other. In this project, I will sort the eigenvectors by their eigenvalues in descending orders, and take the first n vectors to form an eigenspace. To compute eigenvectors, we need to compute co-variance matrix first. Matrix $A$ is the $(k, x)$ sample training data, which contains $k$ samples and each sample has size $x$. The mean vector of A, whose shape is $(1, x)$, is calculated and we denote it $M$. Then, the co-variance matrix $\Sigma$ is calculated by:
\begin{equation}
    \sigma = \frac{1}{N} \sum_k (A_k - M)(A_k - M)^T
\end{equation}
After the covariance matrix is calculated, the eigenvectors and eigenvalues can be computed from the covariance matrix. Then, the eigenvectors are sorted descending by their eigenvalues. The top N eigenvectors are selected to form an eigenspace.
The covariance matrix $\sigma$ can be approximated by a matrix of lower rank. Instead of finding the eigenvectors of the larger system, we can find the eigenvectors of the $k \times k$ system:
\begin{equation}
    A^T Av = \mu v
\end{equation}

\begin{center}
    $AA^TXv = \mu Av$
\end{center}
Therefore, the eigenvectors are $A^TA$, then $Av$ is an eigenvector of $\Sigma$. In order to determine the correct labels of the samples, images need to be projected on the eigenspace by subtracting the mean vector and multiplying the eigenvector matrix. Then, the KNN clustering algorithm is applied to classify the projections on the eigenspace. The predicted labels on the projection are considered the denoising results of the training samples.

\subsubsection{Benchmark 3: Convolutional Autoencoder with KNN Classifier}
To demonstrate that the convolutional autoencoder can lead to enhancement in identifying mislabeled samples, the third benchmark method is introduced. Same as the previous two benchmark methods, a set of representatives that consists of 1,000 samples with labels are selected for classification purposes. A convolutional autoencoder is trained following the identical procedure as the proposed network, defined in Section 4.3, by utilizing the selected representatives as training data. The variation of this benchmark method is that a K-nearest neighbor classifier is applied instead of DBSCAN in the proposed network.

After the autoencoder is obtained, a KNN classifier is fitted by the latent space projections of the representatives. The classifier is utilized to classify the latent projections of all the samples. The classification results are served as the corrections of labels.

\subsection{Results}
Table \ref{tab:1} presents the comparing between the accuracies of different denoising methodologies. To demonstrate the effectiveness of convolutional autoencoder, we can observe the results of three benchmarks. All the three benchmarks are using KNN classifier to reclassify the labels based on a group of selected representatives. Therefore, the results are relevant to the quality of feature preservation in a lower dimensional space.

The comparison indicates that the convolutional autoencoder achieves better performance on feature preservation. Benchmark 1 does not provide any dimension reduction prior to computing the Euclidean distance, all the details in the images are used in the classification, thus the worst performance is obtained. Benchmark 2 and 3 utilizes two approaches to preserving the most significant features and performing classification on the preserved features. The results show that the latent space projections from the convolutional autoencoder are more effective than the independent components extracted from eigenvectors. Therefore, we observed that the latent space data obtained from the convolutional autoencoder does enhance the classification accuracy.

\begin{table}[!htbp]
\caption{Denoising Results}
\begin{center}
 \begin{tabular}{||c c c c||} 
 \hline
 Dataset & MNIST & Fashion MNIST & Cifar 10 \\
 \hline
 Original(Mislabeled) & 0.85 & 0.85 & 0.85 \\
 Benchmark 1 & 0.8437 & 0.7465 & 0.1047 \\
 Benchmark 2 & 0.8963 & 0.7710 & 0.2597 \\
 Benchmark 3 & 0.9194 & 0.8308 & 0.4125 \\
 CAE+DBSCAN & \textbf{0.9591} & \textbf{0.9582} & \textbf{0.8805} \\
 \hline
\end{tabular}
\end{center}
\label{tab:1}
\end{table}

Our network CAE+DBSCAN does achieve the highest accuracy in all three datasets, and the accuracy is significantly improved compared with the three benchmarks. This indicates the great potential of applying DBSCAN on the latent projections encoded by the CAE.

\subsection{Denoising Performance and Reconstruction Quality}
The performance of the model is positively related to the reconstruction quality for all of the three datasets. The reconstruction quality is evaluated by PSNR (Equation 1) and handled by the number of epochs for training the convolutional autoencoder. The denoising accuracy has a stronger positive relationship with the reconstruction quality while the reconstruction quality is low. As the reconstruction quality increases, the improvement of the denoising accuracy becomes slower and eventually approaches a horizontal asymptote. Which the horizontal asymptote is the maximum accuracy the model can reach.

\begin{table}[!htbp]
\caption{Denoising Accuracy vs. Reconstruction Quality}
\begin{center}
 \begin{tabular}{||c c c c c||} 
 \hline
 & & MNIST & &\\
 \hline
 Epochs & 1 & 4 & 7 & 10 \\
 PSNR & 43.15 & 55.82 & 63.29 & 69.03 \\
 Accuracy & 0.9137 & 0.9492 & 0.9568 & \textbf{0.9591} \\
 
 \hline
 & & Fashion-MNIST & &\\
 \hline
 Epochs & 1 & 4 & 7 & 10 \\
 PSNR & 39.83 & 54.83 & 61.23 & 68.78 \\
 Accuracy & 0.8829 & 0.9374 & 0.9512 & \textbf{0.9582} \\
 \hline
 
 & & Cifar 10 & &\\
 \hline
 Epochs & 1 & 4 & 7 & 10 \\
 PSNR & 10.25 & 13.93 & 16.13 & 18.77 \\
 Accuracy & 0.8202 & 0.8593 & 0.8712 & \textbf{0.8805} \\
 \hline
\end{tabular}

\end{center}
\end{table}

\section{Future Research}
The performance of the model presented in this article decreases as more high-resolution images are involved. In real-world settings, the training data are sometimes consisting of details in multiple aspects. The high-complexity, high-resolution mislabeled training samples are harder to be identified by this model.

A future direction of this research will be to extend the autoencoder to a variational autoencoder. A variational autoencoder (VAE) provides a probabilistic manner for describing an observation in latent space. Rather than building an encoder that outputs a single value to describe each latent state attribute, the encoder will describe a probability distribution for each latent attribute. In order to maximize the variance between the probability distributions, some clustering losses may be introduced such as the Soft Normalized Cut Loss\cite{tang2018normalized}.

The Soft Normalizing Clustering(Soft NC) distance \cite{xia2017wnet} of each batch samples is computed:

\begin{equation}
\mathcal{L}_{nc-cluster}:= L - \sum_{l=1}^L \frac{\sum_{i=1}^N a_{il}(\sum_{j=1}^N a_{jl}W_{ij})}{\sum_{i=1}^N a_{il}(\sum_{j=1}^N W_{ij})} \\
    = L - \sum_{l=1}^L \frac{\tilde{\vec{a}}_l^T W\tilde{\vec{a}}_l}{\tilde{\vec{a}}_l^TW\vec{1}}
\label{eq:soft:nc}
\end{equation}

Here $W_{ij}= d(\vec{z}_i,\vec{z}_j)$ are latent variable distance. Since $z_i,z_j$ are stochastic, by reparametrization, this is equivalent to Wasserstein distance:

\begin{equation}
    d(\vec{z_i},\vec{z_j})= \|\vec{\mu}(\vec{x}_i|\theta) - \vec{\mu}(\vec{x}_j|\theta)\|_2^2 + \|\vec{\sigma}(\vec{x}_i|\theta)-\vec{\sigma}(\vec{x}_j|\theta)\|_2^2.   
\end{equation}

The Soft NC Loss can maximize the distances between clusters. Therefore, integrating the Soft NC Loss with the variational autoencoder can possibly maximize the difference between the probability distributions of each class in the latent space. Thus the outliers detected by the DBSCAN algorithm are more likely to be the mislabeled samples.

Another future direction is to utilize a Gaussian Mixture Variational Autoencoder (GMVAE). There are many methods to generate latent clusters. One of the possible approaches solving a marginalized Gaussian Mixture model as the prior distribution of the latent space for identifying latent clusters. Thus, GMVAE can be utilized. The GMVAE is a variant of the variational autoencoder model with a Gaussian mixture as a prior distribution, with the goal of performing unsupervised clustering through deep generative models\cite{dilokthanakul2017deep}. GMVAE \cite{dilokthanakul2016deep} assumes the underlying the latent space random variable $\vec{z}$, when marginalized, is a mixture of Gaussian with $S$ Gaussian kernels:
\vspace{-0.3em}
\begin{align}
y  &\sim p(y), \\
\vec{z}| y=s  & \sim \mathcal{N}(\vec{\mu}_s,diag(\vec{\sigma}_s^2)). \label{eq:z:gmm}
\end{align}
$p(y)$ is the priori distribution of labeling. The conditional distribution of $z$, given the labeling, obeys the Gaussian distribution with mean $\vec{\mu}_s$ and the diagonal covariance $diag(\vec{\sigma}_s)$. The distribution of $z$ is thereby can be stated as:
\vspace{-1em}
\begin{equation}
    p(z) = \sum_{s=1}^S p(z|y=s)p(y=s).
\end{equation}
A further developed GMVAE model called split-GMVAE \cite{charakorn2020explicit} could be a possible implementation that integrated into our network. In split-GMVAE, The prior distribution of $\vec{z}$ and the encoding of $\vec{z}$ generated from $q$ are all set to be multivariate Gaussian. The posterior distribution of soft-labeling vector $\vec{y}$ is generated from the encoder with an additional fully connected layer and a softmax layer.

The GMVAE approach is based on the assumption that the samples from each class are in a Gaussian distribution. Thus the entire dataset with multiple labels (classes) is a mixture of multiple Gaussians. Based on this assumption, there is a Gaussian distribution associated with each class, the data points beyond a specified number of the standard deviations of the specific Gaussian distribution are considered mislabeled data. The data points that are assigned to the wrong Gaussian distribution can be classified as incorrectly labeled samples as well.

\section{Conclusion}
This article presents a straight forward method for classifying incorrectly labeled samples in the training data. This method utilizes convolutional autoencoder for dimension reduction and feature preservation. The results of an empirical evaluation demonstrated that the convolutional autoencoder improves classification accuracy for datasets that possess labeling errors. The method presented can also be optimized and utilized in various types of datasets, not restricted to images. Correcting mislabeled training samples to enhancing the accuracy of the training dataset will significantly affect the results in supervised learning. Therefore, the method presented can be widely applied to remove noised training data in supervised learning.

\section{Acknowledgement}
Thanks to Dr. Andrew Whinston, Professor in McCombs School of Business and the Department of Computer Science-University of Texas at Austin, for the supervision and guidance throughout this project. Also thanks to Dr. Risto Miikkulainen, Professor in the Department of Computer Science-University of Texas at Austin, for very helpful insights and comments regarding machine learning in earlier drafts.

\bibliographystyle{unsrt}
\bibliography{ref}  

\end{document}